\documentclass[conference]{IEEEtran}
\IEEEoverridecommandlockouts
\usepackage{cite}
\usepackage{amsmath,amssymb,amsfonts}
\usepackage{booktabs,subcaption,dcolumn}
\usepackage{algorithmic}
\usepackage{graphicx}
\usepackage{textcomp}
\usepackage{lipsum}
\usepackage{hyperref}
\usepackage{tabularray}
\usepackage{float}
\usepackage{xcolor}
\begin{document}

\title{ A Framework for Feasible Counterfactual Exploration incorporating
Causality, Sparsity and Density\\

}
\author{
\IEEEauthorblockN{Kleopatra Markou}
\IEEEauthorblockA{
\textit{Department of Informatics and}\\
\textit{Telecommunications, National and}\\
\textit{Kapodistrian University of Athens}\\
Athens, Greece \\
klmark@di.uoa.gr}
\and
\IEEEauthorblockN{Dimitrios Tomaras, Vana Kalogeraki}
\IEEEauthorblockA{
\textit{Department of Informatics} \\
\textit{Athens University of Economics and Business}\\
Athens, Greece \\
\{tomaras,vana\}@aueb.gr}
\and
\IEEEauthorblockN{Dimitrios Gunopulos}
\IEEEauthorblockA{
\textit{Department of Informatics and}\\
\textit{Telecommunications, National and}\\
\textit{Kapodistrian University of Athens}\\
Athens, Greece \\
dg@di.uoa.gr}
}

\maketitle

\begin{abstract}
The imminent need to interpret the output of a Machine Learning model with counterfactual (CF) explanations – via small perturbations to the input – has been notable in the research community. Although the variety of CF examples is important, the aspect of them being feasible at the same time, does not necessarily apply in their entirety. This work uses different benchmark datasets to examine through the preservation of the logical causal relations of their attributes, whether CF examples can be generated after a small amount of changes to the original input, be feasible and actually useful to the end-user in a real-world case. To achieve this, we used a black box model as a classifier, to distinguish the desired from the input class and a Variational Autoencoder (VAE) to generate feasible CF examples. As an extension, we also extracted two-dimensional manifolds (one for each dataset) that located the majority of the feasible examples, a representation that adequately distinguished them from infeasible ones. For our experimentation we used three commonly used datasets and we managed to generate feasible and at the same time sparse, CF examples that satisfy all possible predefined causal constraints, by confirming their importance with the attributes in a dataset. 
\end{abstract}


\section{Introduction}

A really important aspect in a Machine Learning (ML) model is the predicted output. Researchers often ask themselves whether the output of a
model can be interpretable and applicable to the real world. In various domains and applications, such as criminal justice \cite{criminaljustice}, clinical healthcare \cite{Clinicalhealthcare}, loan approvals or hiring individuals, a valid explanation provided by the system plays a crucial role to future actions.  In addition,  \cite{Wachter} displays in detail the legal rights that an individual has, regarding an explanation over the output, since its personal data has been used in similar occasions. 

For all the aforementioned reasons, the need to explain these ML models to people and provide them with the insights into how they should act to obtain the desired prediction result, is imminent \cite{Miller}. Counterfactual explanations \cite{Wachter} are often used as a type of local explanations, because they are consistent with the ML model and can be interpretable. As Local Counterfactual (CF) explanation, we consider a representation of the small perturbations of an input feature that are proximal to it and that can lead to a change in the final prediction of the model. The loan example, which is commonly used to describe this definition, answers to the question e.g. “What an  individual should change so the bank will grand him/her the loan that now cannot get?”. All the possible scenarios that alter the class of the original input value are considered as CF examples. An illustrative example of such scenarios is illustrated in the following figure, Figure \ref{fig:cfexampledefinition}. 

\begin{figure}[!ht]
    \centering
    \includegraphics[width=0.5\textwidth]{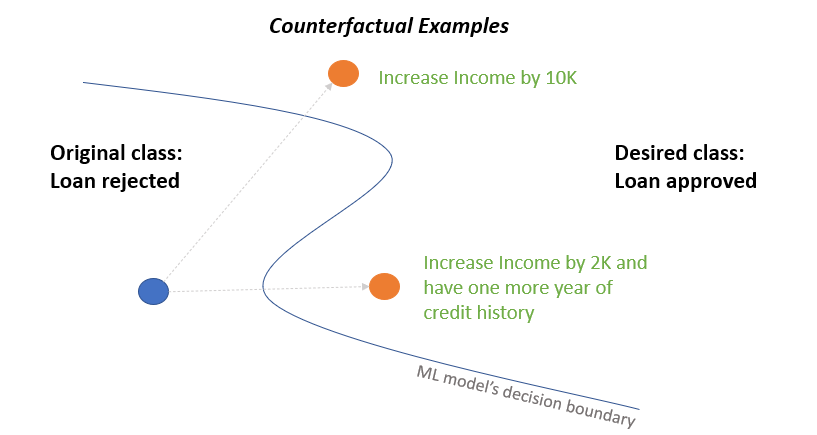}
    \caption{Illustrative example of Counterfactual Explanations}
    \label{fig:cfexampledefinition}
\end{figure}

However, there are several questions arisen regarding these examples. Are actually all these scenarios applicable to the real-world? Are all counterfactual explanation easily adaptable to real-world applications? The answer is negative to these questions. A Machine Learning model can learn from data distributions to recognize patterns and try to imitate them, but it lacks logic. Its produced results cannot be applicable to a multifactorial real-world scenario. Therefore, it is quite common to see counterfactual examples suggesting unachievable goals. For example, a hiring application would propose to a candidate to obtain a new degree in a two month period or in extreme cases become younger to get a particular job. \textit{Feasibility} is a mean for determining this answer, since the scenarios originate from real-world application cases and it can be defined from the causal relations between features through constraints. Given the fact that it is not always possible to reconstruct a full causal graph or a partial one that depicts the dependencies between the features in a dataset, a human can provide domain-knowledge to help with the formulation of logical constraints \cite{Mahajan}. In this work, we propose a simple methodology that can generate counterfactual examples that are able to satisfy causal constraints sufficiently.

\begin{figure}[!ht]
    \centering
    \includegraphics[width=0.5\textwidth]{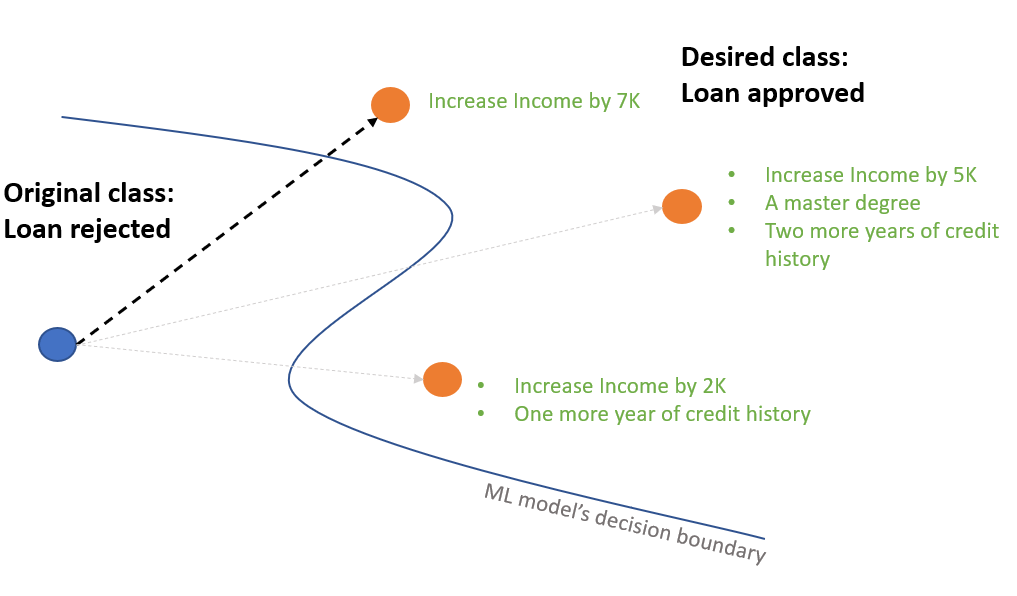}
    \caption{Illustrative example of possible feasible CF examples and sparsity }
    \label{fig:cfexampleandsparsity}
\end{figure}

Counterfactual explanations may provide people with alternative suggestions to follow and by making these explanations compliant with real-world occasions, people can design their own pathways to the solution of their problem. However, the question arising here is after how many changes a solution can be considered as impractical and demanding. It is highly typical that the users would like to follow the fewest amount of changes in order to get the desired output. Therefore, this calls for an appropriate metric to denote that amount of changes and for that reason we added a sparsity formulation in our method to enhance the robustness of our model \cite{Sparsity}. As we can see in Figure \ref{fig:cfexampleandsparsity}, there are three feasible counterfactual examples that suggest three different ways an individual can take a loan. The selected counterfactual example should be the one that suggests the fewer changes, (the orange dot, marked in black dashed line). In this work we are proposing adding sparsity, so the model can learn to generate counterfactual explanations by applying the smaller amount of perturbations.   

\begin{figure}[!ht]
    \centering
    \includegraphics[width=0.5\textwidth]{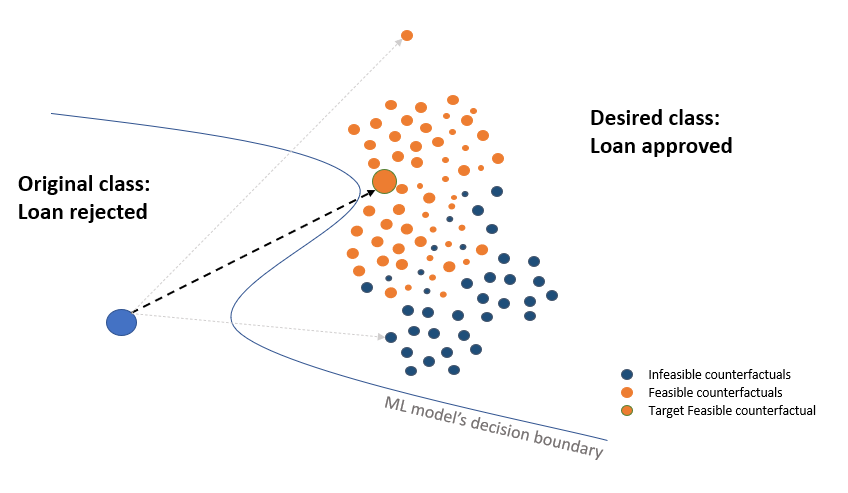}
    \caption{Illustrative example of possible feasible CF examples and manifold representation}
    \label{fig:cfexampleanddensity}
\end{figure}

Apart from the characteristics that a counterfactual example can be labeled with, feasibility, sparsity etc. there are other aspects which determine whether we are going to select a counterfactual or not. In this work we are examining an element that answers the question "where I can locate the examples satisfying all parameters and avoid unrealistic ones?". With the latent space produced by our main model, which we analyze further in Section \ref{sec:architecture}, we are forming manifold representations, i.e. two-dimensional surfaces that represent the various counterfactual examples as real points depicted in these spaces. On these surfaces we can locate dense regions with feasible or infeasible counterfactual examples. As Figure \ref{fig:cfexampleanddensity} shows, there are three feasible counterfactual examples that suggest three different ways an individual can take a loan. The selected counterfactual example should be the one that's closer to the input value and into a dense space with the majority of feasible counterfactual example. The reason for the rejection of the other two examples, is two-fold. First, the counterfactual example that is characterized as infeasible is unlikely to be used in a real-life scenario. Second, the counterfactual example that is far from the dense batch of the other feasible examples, maybe works as an outlier, suggesting for example a much more demanding way of an individual getting a loan. 

Finally, in this paper, our main contributions can be summarized as follows: 
\begin{itemize}
    \item We present the important role of the feasibility in counterfactual examples, through causal relations of the features of each dataset, in the form of constraints that must be satisfied.
    \item We are extending the notion of counterfactual examples feasibility by adding sparsity as a valuable characteristic that helps as to be more efficient and concise, suggesting the fewest amount of changes while generating counterfactual examples.
    \item We exploit the notion of data manifolds and present two-dimensional manifolds that depict the density of feasible and infeasible counterfactual examples in a two-dimensional space from the latent space of the Variational Autoencoder that we are using in this work.
\end{itemize}

\section{Related work}

During the last few years, there has been a variety of works in explainable machine learning (XAI) \cite{bodriaXAI,vermaREVIEW,karimi2021survey,cem,DiCE,REVISE,C-CHVAE,Mahajan,Papapetrou}. As a result, counterfactual examples have attracted a lot of research attention. The work of \cite{vermaREVIEW} has made an extensive review over the algorithmic variability that exists in this field. It also provides us with an overview of research gaps and future ideas to help the continuation of research. A representation of methods on algorithmic recourse (the actions required for “the systematic process of reversing unfavorable decisions by algorithms across a range of counterfactual scenarios”), that work towards providing explanations and recommendations to maltreated individuals is presented in \cite{karimi2021survey}. There are methods that did not use supervised learning techniques but went a step further into using self-supervised ones, such as in the work of \cite{cem}, a contrastive learning method to help justify the classification of inputs by a black box classifier. In addition, there are several works that focus on different properties of counterfactual explanations. For instance, \cite{DiCE} focuses on the diversity of counterfactual explanations and it is also used as constructed a python library to help users generate counterfactual explanations and \cite{REVISE} proposes generating actionable counterfactual examples by applying fewer, more sparse changes. The work of \cite{C-CHVAE} points to the importance of counterfactual explanations via faithfulness, a combination of proximity (a counterfactual should not be local outlier) and connectedness (a counterfactual is close to the correct input value). Finally, the authors of \cite{Mahajan} focus on feasibility, by utilizing structural causal models.  

\section{Methodology}

In this section, we provide an in depth presentation of our problem definition. First, we begin by introducing the theoretical background of our methodology, then we present the components of our model and finally, we describe in detail its final architecture.

For an easier comprehension of the following analysis of the method that we followed, it is crucial to give a formal definition of the notion of counterfactual feasibility. 

{\bf Definition.} (\textit{Feasibility}) : Let the ($x_i$, $y_i$) be the input features and the predicted outcome of the model and $y’$ be the desired output class. We define a counterfactual example ($x_{cf}$, $y_{cf}$) as \textit{feasible} if the desired class $y’$ is equal to the $y_{cf}$ output, the changes from $x$ to $x’$ satisfies all constraints provided by domain knowledge and all variables that conduct a \textit{causal model} (i.e. a structure that depicts all possible relations between the variables of a dataset), lie within the input domain \cite{Mahajan}.

\subsection{Causal constraints}
\label{causalconstraints}

Let us assume that we design constraints which capture feasibility of a counterfactual CF example, using a causal model or even basic domain knowledge, for a certain dataset and its attributes. For example, if  individuals want to take a loan but they need to change certain aspects of their life, a CF example that decreases the attribute “age”, will be considered as infeasible, since it violates the logical causal constraint that age can only increase over time \cite{Mahajan}. 

In order to generalize our approach for any goals and purposes, since it is difficult to create causal graphs depicting the dependencies among various attributes, we based our methodology to logical constraints that can easily be structured with basic domain knowledge. For this reason, we made the first/simpler model, using only one attribute to form the \textit{Unary constraint model} e.g. see Equation (\ref{eq:unarycon}) so that we can define the functionality of our model and in addition we made paired combinations with the attributes that were able to form logical constraints. This model is called \textit{Binary constraint model} e.g. see Equation (\ref{eq:binarycon}).

To give an illustrative example, we utilize one of the datasets presented thoroughly in our experimental evaluation section, Adult dataset\cite{misc_adult_2}, that has the following attributes: \textit{age} and \textit{education}. It is purely logical to assume that if individuals obtain a degree, their age will be higher. In this case, we can formulate the following equations to add to our feasibility evaluation model, so that it can learn to satisfy them:

\begin{equation}
\label{eq:unarycon}
x_{age}^{cf} \geq x_{age}
\end{equation} 

\begin{equation}
\label{eq:binarycon}
\begin{array}{l}
\left(x_{ed}^{cf}>x_{ed} \Longrightarrow x_{age}^{cf}>x_{\text {age }}\right) \text { AND } \\
\left(x_{ed}^{cf}=x_{ed} \Longrightarrow x_{\text {age}}^{cf} \geq x_{\text {age}}\right)
\end{array}
\end{equation}

\subsection{Sparsity}
\label{sec:sparsity}

An important aspect, regarding the effectiveness of a counterfactual explanation, is \textit{sparsity}. In order to generate a counterfactual example, we should change the fewest amount of features/attributes, because end-users tend to follow a smaller list of changes, much easier. We define sparsity as the total number of changes of features that an individual needs to perform to alter its class \cite{Sparsity}. 
    
\subsection{Architecture}
\label{sec:architecture}
\begin{figure}[h]
    \centering
    \includegraphics[width=0.49\textwidth]{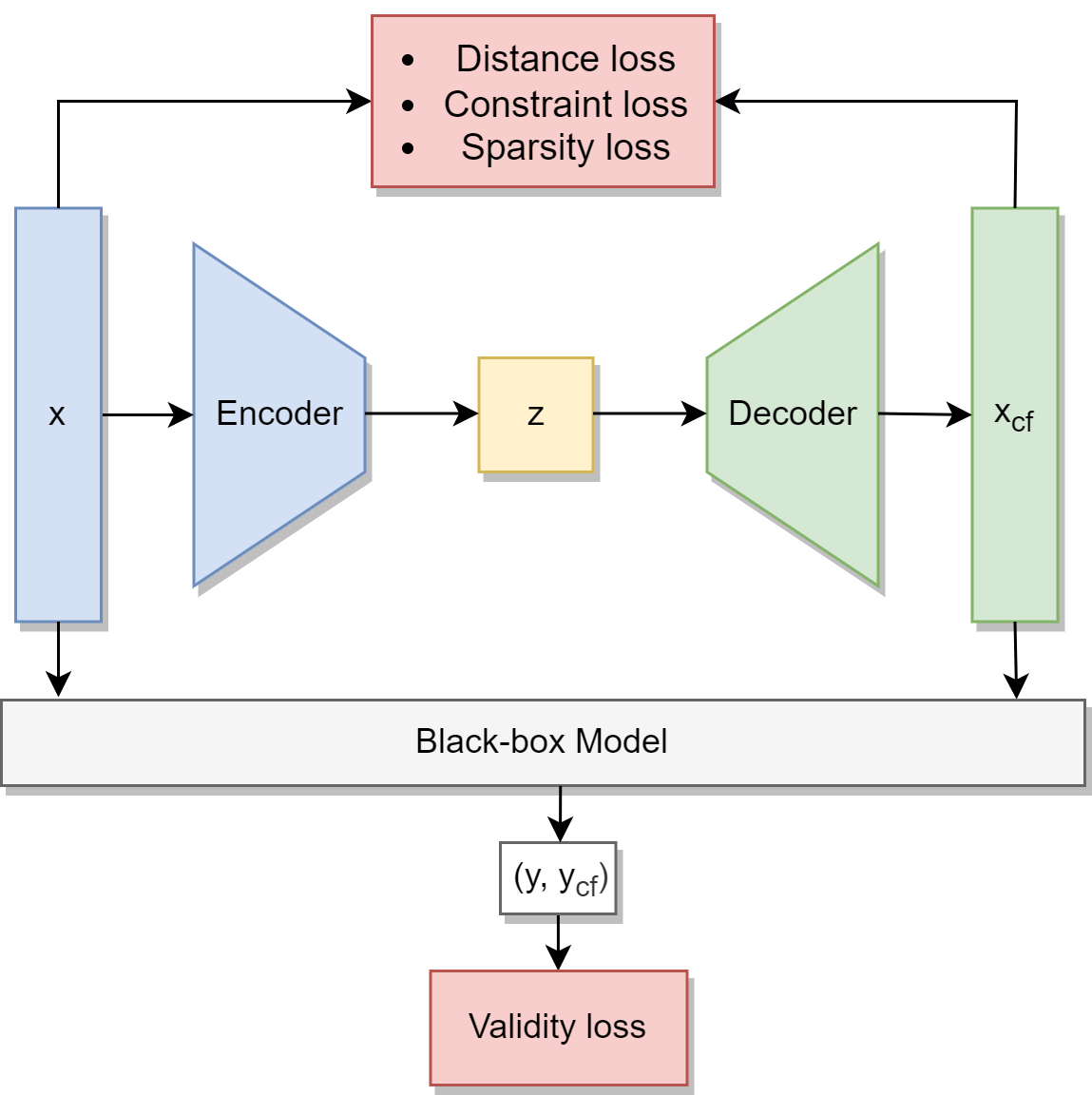}
    \caption{Architecture}
    \label{fig:architecture}
\end{figure}

In this section we present the architecture of our feasibility model, as depicted in Figure \ref{fig:architecture}. During training the input values $X$ are inserted to the encoder of a Variational Autoencoder (VAE) \cite{autoencoding}, where a lower dimensional representation is produced for the latent space. Then the decoder is trying to reconstruct each input value to its original size. Every output result must constitute a counterfactual example. For every epoch, during training, a four-part loss function is applied to train the network to produce counterfactuals that satisfy proximity, validity, feasibility and sparsity. The black-box model is utilized to predict the class of both input and counterfactual examples and at the same time is also used to the validity loss.

{\bf Model Steps.}
At first, we train a black box model, in this case two linear layers, to classify the input data into two classes. This satisfies the definition of a CF example, in which we will always have the input and the desired (opposite) class. The result of this part of the architecture will be later used in the validity loss function as a pretrained model to predict the correct class.

\begin{figure}[H]
    \centering
    \includegraphics[width=0.47\textwidth]{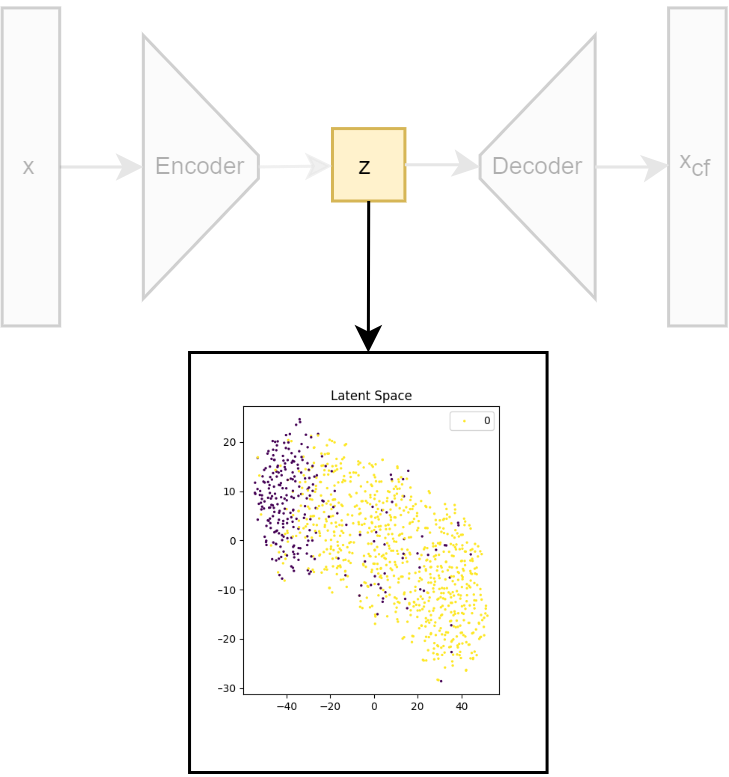}
    \caption{Manifold}
    \label{fig:manifold}
\end{figure}

As a second and final step of our model we are utilizing a Variational Autoencoder (VAE) to generate feasible counterfactual examples. Our main goal is to generate approximate CF examples to the input data and at the same time they should be feasible by satisfying the causal constraints. For this reason, the encoder of the VAE, which works as a bottleneck, is trained with heterogeneous data $X$ and produces a low dimensional representation of them $z$. This representation is really useful because it can approximate the density of data in a lower dimensional space, by giving us a manifold (see Figure \ref{fig:manifold}). Since we are aiming to generate counterfactuals, perturbed representations of the input samples, we perturbed the output of the encoder to the decoder. The decoder must conduct a faithful representation of the input data. Its output finally passes through the pre-trained black-box model, which we mentioned earlier at the beginning of Section \ref{sec:architecture}, so that we can examine the prediction. An abstract representation of our model is depicted in Figure \ref{fig:architecture}.

{\bf Immutable Attributes.}
An important aspect, during the training of the VAE, which can influence its outcome, has to do with the handling of the immutable attributes. We consider an attribute immutable, when it should remain the same through the training of the model \cite{REVISE}. For example we consider the attribute \textit{race} immutable because an individual cannot changed its race, even if the counterfactual explanation suggested such change. Such attributes were disabled for the training of the VAE model but we incorporated them again in the final prediction of our model.

{\bf Loss function.}
For the training of the VAE to be able to generate accurate and feasible counterfactual examples, we created a loss function that includes the main components from the CF definition and in addition the feasibility constraints and sparsity. In total, four elements should be satisfied, proximal distance, desired class (validity) , feasibility constraints and sparsity, otherwise our model is penalized. 

In Equation (\ref{cfdefinitionloss}) the first part the validity of the counterfactual is calculated with HingeLoss, from the comparison between the $h(x_{cf})=y_{cf}$, the predicted class of the CF example and $y'$, the desired class (opposite class from the output of the classifier $y$). The second term measures with the $L1$ the distance between the counterfactual example and the input value. 

\begin{equation}
\label{cfdefinitionloss}
\operatorname{argmin}_{x^{cf}} \operatorname{Loss}\left(h\left(x^{cf}\right), y^{\prime}\right) + d(x,x')
\end{equation}

As for the two types of constraints, we are adding for the unary constraint the $-min(0,x^{cf}-x)$ term and $+(x_2 - {c_1} - {c_2}x_1)-min(0,{c_2})$ for the binary, where $x_1$, $x_2$ are the two features that form the logical constraint and $c_1$, $c_2$ are parameters selected from experimentation. 

The final part of the loss function $+g(x'-x)$, denotes sparsity between the input value and the generated counterfactual example from their features, calculated with $L0/L1$ norm.

\section{Experiments and Results}

In this section, we provide an extended description of the datasets that we used, the preprocessing techniques that applied to them and in Subsection \ref{eval_metrics} we fully present all metrics that have been utilized to evaluate the model. In addition, an in depth overview of the experimental results of our method against a variety of different methods that generate counterfactual CF examples is presented for all tree datasets. 

\subsection{Datasets}
We use three publicly available datasets as benchmarks. Our choice is based on their established and constant use in many previous works. A description will follow. In addition, Table \ref{table:DatasetanOverview} summarizes the datasets’ information, including the total number of instances, the number of categorical, binary and continuous attributes, the final count of inputs that remained after the preprocessing and the selected decision/target class. Finally we followed a 80\%:10\%:10\% split of the dataset for training, validation and testing, respectively. 

\begin{table}[htbp]
\caption{Datasets: an overview}
\setlength{\tabcolsep}{0.5\tabcolsep}
\begin{center}
\begin{tabular}{c|cccc}
\hline \multicolumn{1}{c|}{ \textbf{Datasets} } & \multicolumn{1}{c}{ \textbf{\# Instances} } & \begin{tabular}{c} 
\textbf{\# Instances}\\
\textbf{(cleaned)}
\end{tabular} & \textbf{\# Attributes$^{\mathrm{*}}$}  & \textbf{Target class} \\
\hline \begin{tabular}{l} Adult \end{tabular} & 48.842 & 32.561 & $5 / 2 / 2$ & Income \\
\begin{tabular}{l} 
KDD-Census \\
Income
\end{tabular} & 299.285 & 199.522 & $32 / 2 / 7$ & Income \\
 \begin{tabular}{l} 
Law School \\
Dataset
\end{tabular} & 20.798 & 20.512 & $1 / 3 / 6$ & Pass the bar \\
\hline \multicolumn{5}{l}{$^{\mathrm{*}}$Number of Categorical/Binary/Numerical attributes.}
\end{tabular}
\label{table:DatasetanOverview}
\end{center}
\end{table}

The \textbf{Adult Income} dataset \cite{misc_adult_2} is a real-world dataset provided by UCI machine learning repository. It is commonly used in similar applications to ours for classifying whether an individual is likely to have an income of more than 50k per year. It includes 48.842 instances in total, from which we used 32.561. We used 9 attributes in total, 5 categorical, 2 binary and 2 numerical/continuous, respectively. As immutable attributes we selected \textit{race} and \textit{gender}.  

The \textbf{KDD Census-Income} dataset \cite{misc_census-income_(kdd)_117} contains weighted census data extracted from the 1994 and 1995 Current Population Surveys conducted by the U.S. Census Bureau. It is used as the Adult to classify whether an individual is likely to have an income of more than 50k annually. This dataset contains 299.285 instances, from which we used 199.522. We used 41 attributes in total, 32 categorical, 2 binary and 7 numerical/continuous, respectively. For this dataset we kept the same immutable attributes as in Adult \textit{race} and \textit{gender}. 

The \textbf{Law school} dataset \cite{Wightman1998LSACNL} contains the law school admission records from 163 law schools in the US in 1991. In this dataset the target is to predict whether a candidate would pass the bar exam or not. It includes 20.798  instances in total, from which we used 20.512. We used 10 attributes in total, 1 categorical, 3 binary and 6 numerical/continuous, respectively. For this dataset \textit{sex} is the immutable attribute.  

\subsection{Implementation Settings}
\label{sub:implset}

In this part of our work, a detailed description of the implementation of the Variational Autoencoder (VAE) is presented in Table \ref{table:ImplVAE}. Both the encoder and the decoder are consisted of five linear layers and a ReLU activation function. We added a dropout of 30\% to each one of these layers. With the \textit{Num. Features} we describe the size of each input coming through the encoder and the target class. The size \textit{Latent space vector} is adjusted to 10 features. 

\begin{table}[H]
\begin{center}
\caption{VAE's implementation settings}
\setlength{\tabcolsep}{0.3\tabcolsep}
\begin{tabular}{l|l|ccc}
\hline \textbf{Encoder}&\textbf{Layers} & \textbf{Input} & \textbf{Output} & \textbf{Activation} \\
\hline & L1 & Num. Features + 1 & 20 & ReLU \\
 &L2 & 20 & 16 & ReLU\\
 &L3 & 16 & 14 & ReLU\\
 &L4 & 14 & 12 & ReLU\\
 &L5 + Sigmoid & 12 & Latent space vec. & ReLU\\
 \hline \textbf{Decoder} & L1 & Latent space vec. + 1  & 12 & ReLU \\
 &L2 & 12 & 14 & ReLU\\
 &L3 & 14 & 16 & ReLU\\
 &L4 & 16 & 18 & ReLU\\
 &L5 + Sigmoid & 20 & Num. Features. & ReLU\\
\hline
\end{tabular}
\label{table:ImplVAE}
\end{center}
\end{table}

In Table \ref{table:ImplDatasets} an overview of the hyperparameter tunning is presented for each one of the datasets that we employed. 

\begin{table}[htbp]
\begin{center}
\caption{Implementation Settings}
\setlength{\tabcolsep}{1\tabcolsep}
\begin{tabular}{llccc}
\hline \textbf{Datasets} & \textbf{Method} & \textbf{Learning rate} & \textbf{Batch size} & \textbf{Epochs}\\
\hline Adult & Unary-const & $0.2$ & 2048 & 25\\
  & Binary-const & $0.2$ & 2048 & 50\\
\hline KDD-Census & Unary-const & $0.1$& 2048 & 25\\
  & Binary-const & $0.1$ & 2048 & 25\\
\hline Law School & Unary-const & $0.2$ & 2048 & 25\\
  & Binary-const & $0.2$ & 2048 & 50\\
\hline
\end{tabular}
\label{table:ImplDatasets}
\end{center}
\end{table}

\subsection{Preprocessing}\label{preprocess}
For the preprocessing, we followed the same techniques, for each dataset. First, all the rows with missing values were deleted from the final format of the datasets. The continuous features were normalized between 0 and 1 and categorical features were converted using one-hot encodings. Finally, we transformed the binary attributes to 0 and 1. The number of the remaining instances is noted in Table \ref{table:DatasetanOverview}.

\subsection{Evaluation Metrics}
\label{eval_metrics}

In this section, we present the evaluation metrics that we used to evaluate the performance of our model regarding the feasibility of counterfactual examples. We compared our results with other methodologies regarding the \textit{i) validity}, \textit{ii) feasibility score}, \textit{iii) continuous proximity}, \textit{iv) categorical proximity} and  \textit{v) sparsity}.

\begin{itemize}
    \item \textbf{\textit{Validity:}} as "Validity", we measure the percentage (\%) of counterfactual CF examples for which the predicted class aligns with the target class from the classifier. 
    \item \textbf{\textit{Feasibility Score:}} as "Feasibility score", we measure the percentage (\%) of the counterfactual CF examples that satisfied the logical constraints (unary or binary) given in the Equations (\ref{eq:unarycon}) and (\ref{eq:binarycon}) in Section \ref{causalconstraints}.
    \item \textbf{\textit{Continuous proximity:}} We first define the average $l_1$ distance between the continuous features of the input value and the corresponding ones of the generated counterfactual CF example and then we finally calculate the mean as the "Continuous proximity", see Equation (\ref{continuous}). The $(-1)$ multiplication, clarifies the proximity term.  

\begin{equation}
-\frac{1}{k} \sum_{i=1}^k \text { dist\_cont}\left(c_i, \boldsymbol{x}\right) \label{continuous}
\end{equation}

    where $k$ is the total number of CF examples and $c_i$ is a certain CF example. 
    
    \item \textbf{\textit{Categorical proximity:}} We initially calculate the total number of alterations of the input value and the corresponding ones of the generated counterfactual CF examples and then we measure the mean as the "Categorical proximity", see Equation (\ref{categorical}). As mentioned above, The $(-1)$ multiplication, clarifies the proximity term.  

\begin{equation}
-\frac{1}{k} \sum_{i=1}^k \text { dist\_cat}\left(c_i, \boldsymbol{x}\right) \label{categorical}
\end{equation}

     where $k$ is the total number of CF examples and $c_i$ is a CF example. 
    
    \item \textbf{\textit{Sparsity:}} We measure the number of features that changes for every input value to its corresponding counterfactual CF example and from them, we calculate the mean "Sparsity score".  
\end{itemize}

\subsection{Results}
We are beginning this Subsection by describing some additional final elements that led us to the extraction of the experimental results.

At first is important to point out that regarding the datasets, we used the \textit{age} attribute and a combination of it with \textit{education}, to form the unary and binary constraint, respectively, for the Adult and KDD-Census datasets, with the attribute \textit{Income} being the target/desired class. As for the Law school dataset, the \textit{lsat} attribute formed the unary constraint and combined with \textit{tier}, we formed the binary one. In this case, \textit{Pass the bar} attribute is the target/desired class.

We present the results from every method that we used for all three datasets in Table \ref{table:ResultsDatasets}. The methods REVISE \cite{REVISE}, C-CHVAE \cite{C-CHVAE}, CEM \cite{cem} and FACE \cite{FACE} were reproduced from a python library CARLA \cite{carla}. CARLA\footnote{\url{https://carla-counterfactual-and-recourse-library.readthedocs.io/en/latest/}} is a library used to benchmark counterfactual explanations that supports Machine Learning (ML) models and a variety of datasets. For the experiments with DiCE \cite{DiCE}, we selected one of its included models, namely $random$. DiCE\footnote{\url{https://www.microsoft.com/en-us/research/project/dice/overview/}} is a python library as well, that helps us explain the predictions of a ML-based system via generating counterfactual explanations. For all these methods feasibility was measured as an evaluation metric using the two logical constraints on the generated counterfactual examples. As for \cite{Mahajan} and our model we trained the two different models separately, one for the \textit{Unary constraint} and the other for \textit{Binary constraint} model, where feasibility was utilized both as a learning parameter and as an evaluation metric. 


From Table \ref{table:ResultsDatasets}, several conclusions can be made regarding the results. At first in two out of three datasets we manage to achieve the highest feasibility scores, in Adult $72.38\%$ and $77.54\%$ and in Law school $93.33\%$ and $86.66\%$, for the Unary constraint model and the Binary one, respectively. In the third dataset, the Census - Income we may have not achieve the best feasibility score possible, although our validity score is $100\%$, in addition to C-CHVAE method with $48.44\%$ for the Unary model and CEM method $86.68\%$. This means that our classifier was better trained and is more likely to give us correct counterfactual examples. Following up, we have validity almost always at $100\%$. From this result we can verify that our model can produce with high probability valid counterfactual examples which have the desired class. Furthermore, we have sparsity, thus we are losing from CEM method in all three datasets. This result does not concern us because it may look to find the minimal amount of changes but it does not achieve our goal. It lacks in validity and feasibility altogether. Finally, both our models, the unary and binary one in the majority of cases reached higher validity, feasibility and sparsity scores against \cite{Mahajan}. Therefore, two main general conclusions have been made: the first is that our methods can be characterized successful since they reduce sparsity to achieve high feasibility scores and validity, and second, the datasets played an important role regarding the final outcome of every method used in this work. 

\begin{table*}[htbp]
    \centering
    \caption{Results }
    \begin{subtable}{1\linewidth}
        \caption{Adult Income dataset}
\begin{center}
\begin{tabular}{c|cccccc}
\hline \multicolumn{1}{c|}{ \textbf{Methods} } & \multicolumn{1}{c}{ \textbf{Validity} } & \begin{tabular}{c} 
\textbf{Feasibility/}\\
\textbf{Unary const.}
\end{tabular} & \begin{tabular}{c}\textbf{Feasibility/}\\ \textbf{Binary const.} \end{tabular} & \begin{tabular}{c}\textbf{Continuous} \\ \textbf{proximity}\end{tabular}  & \begin{tabular}{c}\textbf{Categorical} \\ \textbf{proximity}\end{tabular} & \multicolumn{1}{c}{ \textbf{Sparsity} } \\
\hline  Mahajan et al.\cite{Mahajan} Unary & 100 & 62.26 & - & -2.40 & -3.29 & 5.32\\
  Mahajan et al. \cite{Mahajan} Binary  & 100 & - & 73.18 & -2.53 & -3.25 & 5.25\\
  REVISE \cite{REVISE} & 54.07 & 7.28 & 6.34 & -5.32 & -3.23 & 5.48\\
  C-CHVAE \cite{C-CHVAE}& 85.15 & 32.50 & 31.75 & -4.66 &-3.66 & 5.9\\
  CEM \cite{cem}& 74 & 21.88 & 21.80 & -13.69 & -0.50 & \textbf{2.10}\\
  DiCE random \cite{DiCE} & 93.82 & 11.65 & 11.52 & -3.83 & -0.64 &3.47 \\
  FACE \cite{FACE} & 83.2 & 32.12 & 30.82 & -4.82 & -3.29 & 5.52\\
\hline \textbf{Our method (a)$^{\mathrm{*}}$} & 98 & \textbf{72.38} & - & -2.38 & -2.66 & 4.33\\
\textbf{Our method (b)$^{\mathrm{**}}$} & 100 & - & \textbf{77.54} & -2.80 & -2.39 & 4.55\\
\hline \multicolumn{7}{l}{$^{\mathrm{*}}$ Unary Constraint model / $^{\mathrm{**}}$ Binary Constraint model}
\end{tabular}
\label{table:ResultsDatasets}
\end{center}
    \end{subtable}
    \hspace{1cm}
    
    \begin{subtable}{1\linewidth}
        \caption{KDD-Census Income dataset}
\begin{center}
\begin{tabular}{c|cccccc}
\hline \multicolumn{1}{c|}{ \textbf{Methods} } & \multicolumn{1}{c}{ \textbf{Validity} } & \begin{tabular}{c} 
\textbf{Feasibility/}\\
\textbf{Unary const.}
\end{tabular} & \begin{tabular}{c}\textbf{Feasibility/}\\ \textbf{Binary const.} \end{tabular} & \begin{tabular}{c}\textbf{Continuous} \\ \textbf{proximity}\end{tabular}  & \begin{tabular}{c}\textbf{Categorical} \\ \textbf{proximity}\end{tabular} & \multicolumn{1}{c}{ \textbf{Sparsity} } \\
\hline  Mahajan et al.\cite{Mahajan} Unary & 100 & 90.08 & - & -3.47 & -7.45 & 9.54\\
  Mahajan et al. \cite{Mahajan} Binary  & 100 & - & 79.62 & -3.47 & -7.45 & 9.53\\
  REVISE \cite{REVISE} & 28.09 & 91 & 75.43 & -7.39 & -11.27 & 20.30\\
  C-CHVAE \cite{C-CHVAE}& 48.44 & \textbf{98.87} & 76.61 & -7.66 &-11.94 & 19.74\\
  CEM \cite{cem}& 86.68 & 86.98 & \textbf{85.88} & -8.90 & -0.45 & \textbf{0.51} \\
  DiCE random \cite{DiCE} & 100 & 93.50 & 88.36 & -4.24 & -1.58 & 9.41\\
  FACE \cite{FACE} & 70.18 & 73.24 & 71.01 & -8.11 & -8.30 & 12.38\\
\hline \textbf{Our method (a)$^{\mathrm{*}}$} & 100 & 94.10 & - & -1.57 & -6.03 & 8.15\\
\textbf{Our method (b)$^{\mathrm{**}}$} & 100 & - & 80.84 & -3.34 & -9.39 & 8.65\\
\hline \multicolumn{7}{l}{$^{\mathrm{*}}$ Unary Constraint model / $^{\mathrm{**}}$ Binary Constraint model}
\end{tabular}
\label{table:ResultsDatasets}
\end{center}
    \end{subtable}
    \hspace{1cm}
    
    \begin{subtable}{0.9\linewidth}
        \caption{Law School Dataset}
\begin{center}
\begin{tabular}{c|cccccc}
\hline \multicolumn{1}{c|}{ \textbf{Methods} } & \multicolumn{1}{c}{ \textbf{Validity} } & \begin{tabular}{c} 
\textbf{Feasibility/}\\
\textbf{Unary const.}
\end{tabular} & \begin{tabular}{c}\textbf{Feasibility/}\\ \textbf{Binary const.} \end{tabular} & \begin{tabular}{c}\textbf{Continuous} \\ \textbf{proximity}\end{tabular}  & \begin{tabular}{c}\textbf{Categorical} \\ \textbf{proximity}\end{tabular} & \multicolumn{1}{c}{ \textbf{Sparsity} } \\
\hline  Mahajan et al.\cite{Mahajan} Unary & 100 & 86.66 & - & -10.53 & -2.26 & 7.8\\
  Mahajan et al. \cite{Mahajan} Binary  & 100 & - & 73.33 & -13.09 & -2.26 & 7.93\\
  REVISE \cite{REVISE} & 100 & 71.87 & 69.51 & -19.55 & -2.23 & 7.81\\
  C-CHVAE \cite{C-CHVAE}& 100 & 73.43 & 70.86 & -11.77 & -2.17 & 8.03\\
  CEM \cite{cem}& 85 & 56.38 & 55.25 & -5.01 & -1.52 & \textbf{2.68}\\
  DiCE random \cite{DiCE} & 55 & 85.48 & 24.24 & -4.68 & -1.16 & 5.64\\
  FACE \cite{FACE} & 100 & 78.12 & 70.67 & -9.54 & -1.95 & 9.3\\
\hline \textbf{Our method (a)$^{\mathrm{*}}$} & 100 & \textbf{93.33} & - & -9.72 & -1.93 & 7.2\\
\textbf{Our method (b)$^{\mathrm{**}}$} & 100 & - & \textbf{86.66} & -12.01 & -2.10 & 7.46\\
\hline \multicolumn{7}{l}{$^{\mathrm{*}}$ Unary Constraint model / $^{\mathrm{**}}$ Binary Constraint model}
\end{tabular}
\label{table:ResultsDatasets}
\end{center}
    \end{subtable}
    \label{table:ResultsDatasets}
\end{table*}

\begin{figure}
     \centering
     \begin{subfigure}[b]{1\linewidth}
         \centering
         \includegraphics[width=1\linewidth]{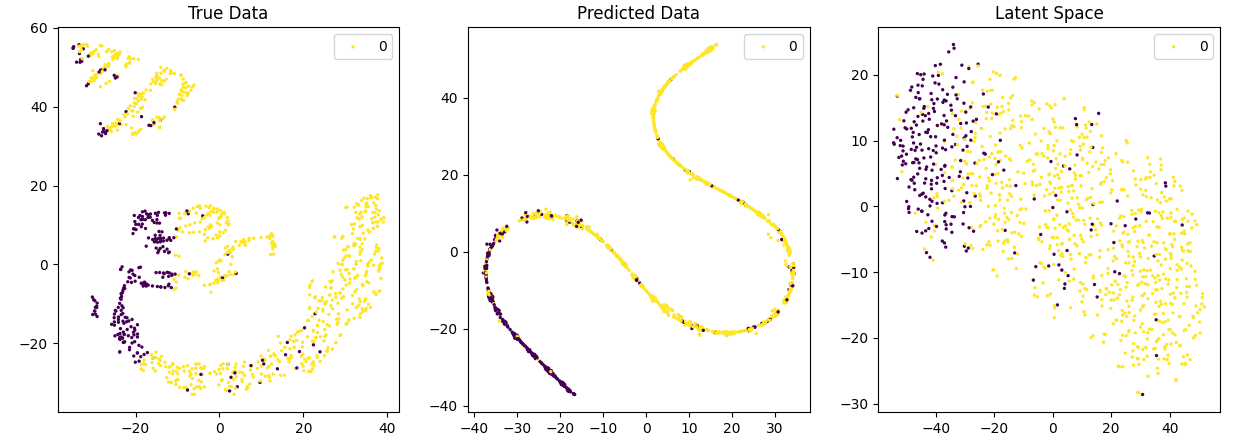}
         \caption{Adult Income}
         \label{fig:tsnemanifoldadult}
     \end{subfigure}
     \hfill
     
     \begin{subfigure}[b]{1\linewidth}
         \centering
         \includegraphics[width=1\linewidth]{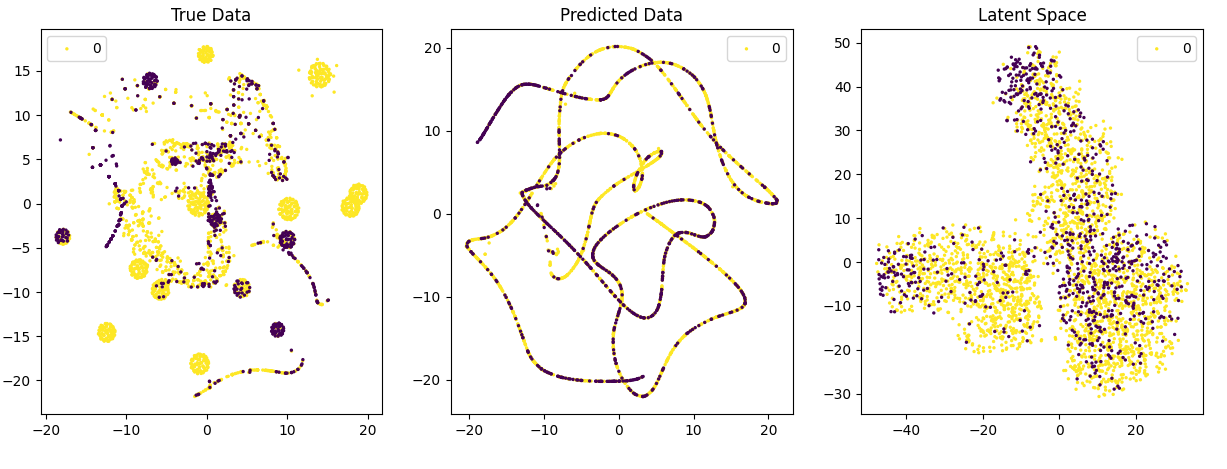}
         \caption{KDD-Census Income }
         \label{fig:tsnemanifoldkdd}
     \end{subfigure}
     \hfill
     
     \begin{subfigure}[b]{1\linewidth}
         \centering
         \includegraphics[width=1\linewidth]{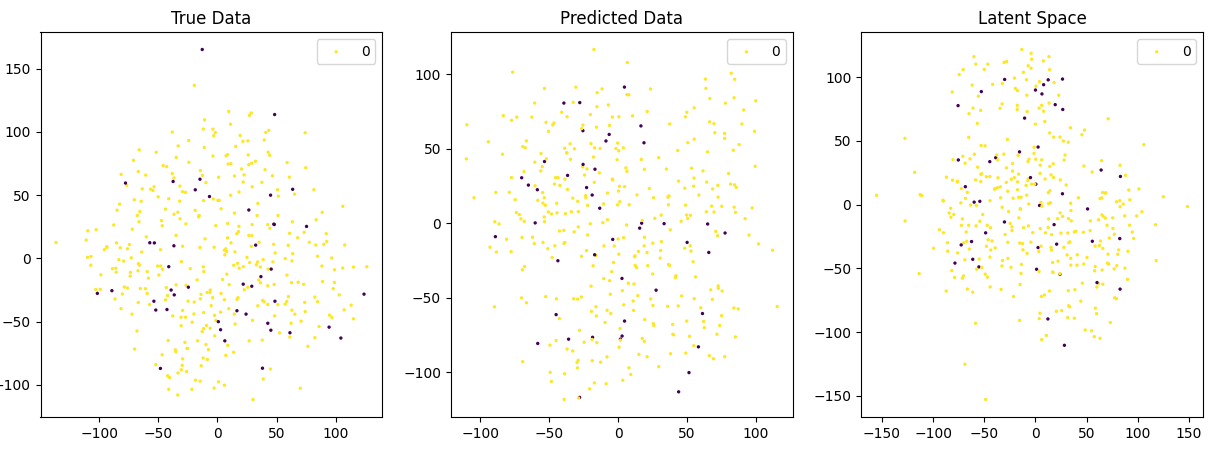}
         \caption{Law School}
         \label{fig:tsnemanifoldlaw}
     \end{subfigure}
        \caption{Datasets' Manifolds. Violet color corresponds to infeasible cf examples and yellow to the feasible ones,}
        \label{fig:manifoldsdata}
\end{figure}

For the extraction of the manifolds for each dataset as depicted in Figure \ref{fig:manifoldsdata} we used the t-SNE by \cite{Hintontsne}, a visualization technique, which represents high-dimensional data by giving each datapoint a location in a two dimensional map. In our case we directly took and saved a sample point from the latent space of the VAE, then passed the points through the decoder and saved the counterfactual example. Consequently, we assigned a label $0/1$ to the cf examples when we had a feasible or infeasible cf example, respectively. Finally, we applied the t-SNE to project the high dimensional latent space vectors in two dimensional representation for the training data, the sample points from the latent space of the VAE and the predicted data, as shown in the three continual diagrams. 
Regarding Figure \ref{fig:manifoldsdata}, we can identify, with some certainty, separable regions in both training, latent space and predicted values diagrams.

Last, in Table \ref{table:SuccessfulCFexample} a successful counterfactual example is presented. The elements marked in red denote the satisfaction of both causal constraints, regarding the relation between age and education. This particular example generated by the binary constraint model from the Adult dataset.

\begin{table}[h]
\begin{center}
\caption{Successful CF example - Adult dataset}
\setlength{\tabcolsep}{1\tabcolsep}
\begin{tabular}{l|cc}
\hline \textbf{Features} & \textbf{x\_true (= our\ input)} & \textbf{x\_pred} \\
\hline age & \textbf{38} & \textbf{43.55} \\
 hours\_per\_week & 40 & 40.36 \\
 workclass & private & private \\
 education & \textbf{hs\_grad}& \textbf{doctorate} \\
 marital status & single & married \\
 occupation & professional & white\_collar \\
 race & white & white \\
 gender & male & male \\
\hline
\end{tabular}
\label{table:SuccessfulCFexample}
\end{center}
\end{table}

\section{Conclusions and Future work}

After a variety of experiments with three vastly known datasets we made several conclusions concerning the results. First, we outperformed the majority of the other methods regarding feasibility. This highlights the significance of this particular property and points to the fact that, feasibility does not always comes along with every method and does not adapt easily to assist real-world scenarios. Another important conclusion was derived from the use of sparsity as a property of our model. From the extraction of the results we can observe that our methods exceed expectations. In most experiments we managed to achieve good results, regarding the changes required for the generated counterfactual examples. The use of immutable values lead partially to this result. With regards to the extracted manifolds, the results were also encouraging. In addition to the previous observations, we confirmed the importance of data structures/dataset in respective applications. The variability, diversity and the quality of data can reach out to better results and feasibility can easily be affected from them. As future work we have already started working on analysing the causal relations of various features in a dataset, so that we can minimize the human involvement during the construction of the causal constraint that will satisfy feasibility.

\section*{Acknowledgment}

This work was supported by the European Union’s Horizon Europe research and innovation programme under grant agreement No. 101070568 (Human-Compatible Artificial Intelligence with Guarantees (AutoFair)).

\bibliographystyle{IEEEtran}

\bibliography{bibliography}{}


\end{document}